\documentclass{article}

\PassOptionsToPackage{numbers,compress}{natbib}
\usepackage[sglblindworkshop, final]{neurips_2026}
\workshoptitle{Machine Learning for Systems @ NeurIPS 2026}

\usepackage[utf8]{inputenc}
\usepackage[T1]{fontenc}
\usepackage{hyperref}
\usepackage{url}
\usepackage{booktabs}
\usepackage{amsfonts}
\usepackage{amsmath}
\usepackage{nicefrac}
\usepackage{microtype}
\usepackage{graphicx}
\usepackage{xcolor}

\newcommand{\tf}{\textsc{full}}

\newcommand{\zonly}{\textsc{zeroonly}}

\title{Tail-Replay: Escaping the Curse of Linear Attention in Prefix Caching for Hybrid LLMs}

\author{%
  Yirui Liu$^{1,*,\dagger}$ \quad Ruoling Qi$^{1,2,*}$ \quad Xuaner Wu$^{1}$ \quad Penghang Liu \quad Jian Chen\\
  $^{1}$Institute of Artificial Intelligence, China Telecom (TeleAI) \\
  $^{2}$Shanghai Jiao Tong University  \\
  $^{*}$Equal contribution \quad $^{\dagger}$Corresponding author \\
  Yirui Liu: \texttt{yiruiliu926@gmail.com} \quad Ruoling Qi: \texttt{qiruoling760@sjtu.edu.cn}
}

\begin{document}

\maketitle

\begin{abstract}
Hybrid large language models interleave full-attention layers with linear-attention layers to reduce the cost of long-context inference. This structure complicates prefix caching: full-attention key-value caches are token-addressable, whereas linear-attention layers maintain recurrent states that cannot be rolled back to arbitrary prefix boundaries. Existing hybrid prefix caching methods address this mismatch by storing recurrent-state checkpoints. As a result, token-level matches are directly usable only at positions aligned with stored checkpoints, constraining prefix reuse to a discrete set of boundaries. We present Tail-Replay, a prefix caching mechanism that enables unconstrained token-level prefix reuse in hybrid large language models. The key insight is that linear-attention mechanisms such as Gated DeltaNet can be viewed as a structured, lossy compression of the input prefix: gated recurrent updates progressively attenuate the contributions of earlier inputs. Consequently, the recurrent state of a matched prefix can be well approximated by replaying only a short, recent suffix of that prefix. Tail-Replay exploits this property by caching the exact full-attention key-value cache while omitting recurrent-state checkpoints. On a cache hit, it reconstructs the linear-attention states by replaying a short, recent suffix of the matched prefix.  As a result, the reuse boundary is determined by the shared tokens rather than by recurrent-state checkpoints.  We evaluate Tail-Replay on three Gated DeltaNet-based hybrid models using the LongBench and RULER benchmarks.  With only a 5--10\% replay budget, it retains 92.8--99.9\% of full-prefill quality on LongBench and RULER. For serving efficiency, we evaluate time-to-first-token speedups across multiple matched-prefix lengths---8K, 16K, and 32K. The speedup grows with prefix length, reaching $9.1$--$14.3\times$ over full prefill at 32K.
\end{abstract}

\section{Introduction}

LLM applications have evolved from single-turn instruction-following assistants~\cite{instructgpt} to richer workloads such as multi-turn dialogue~\cite{lmsyschat}, retrieval-augmented generation~\cite{rag}, tool-using agents~\cite{react}, and multi-agent workflows~\cite{autogen,metagpt}. These workloads bring longer contexts, repeated invocations, and higher concurrency, making serving efficiency an increasingly important practical concern~\cite{mooncake}. In response, model designers and system designers have pursued complementary solutions at different layers of the stack. At the model level, hybrid architectures~\cite{jamba,samba,nemotronh,minimax01,qwen35,qwen36,glm53flash,olmohybrid} combine full-attention (FA) layers with linear-attention layers to reduce long-context cost. At the serving level, prefix caching~\cite{sglang,promptcache,cachedattention,cacheblend} reuses shared prefixes across requests to avoid redundant prefill.

Hybrid models and prefix caching are two complementary techniques. A natural question is whether they can be deployed together to achieve greater overall efficiency gains. However, existing prefix caching mechanisms are largely designed for FA models and are not naturally compatible with hybrid models. In FA models, the reusable state consists of token-indexed key-value (KV) caches, so the prefix KV of a processed request can be retrieved and reused at any token boundary. In contrast, linear-attention layers summarize the processed prefix into recurrent states through in-place updates. Once the state has advanced, it cannot be rolled back to represent an arbitrary earlier prefix. We refer to this mismatch as the \textbf{curse of linear attention for prefix caching: a token-level prefix match no longer directly implies a reusable model state}. Recent hybrid prefix caching systems address this constraint by explicitly managing recurrent-state checkpoints. Marconi~\cite{marconi} focuses on which recurrent states to retain across cached prefixes, while Sparse Prefix Caching~\cite{sparseprefix} focuses on where to place checkpoints within each cached prefix. These designs mitigate the problem but do not remove its root limitation: prefix reuse remains constrained by recurrent-state checkpoint locations rather than by the boundaries of shared-token prefixes.

We present Tail-Replay, a prefix caching mechanism that enables unconstrained token-level prefix reuse in hybrid LLMs. The key insight is that state-of-the-art linear-attention mechanisms such as Gated DeltaNet (GDN) can be viewed as a structured, lossy compression of the input prefix: gated recurrent updates progressively attenuate the contributions of earlier inputs. Consequently, the recurrent state of a matched prefix can be well approximated by replaying only a short, recent suffix of that prefix. Tail-Replay exploits this property by caching the exact FA KV while omitting recurrent-state checkpoints. On a cache hit, it reconstructs the linear-attention states by replaying a short, recent suffix of the matched prefix. As a result, the reuse boundary is determined by the shared tokens rather than by recurrent-state checkpoints, escaping the curse of linear attention in prefix caching.

We evaluate Tail-Replay on three Gated DeltaNet-based hybrid LLMs across LongBench and RULER. With a 5--10\% replay budget, it retains 92.8--99.9\% of full-prefill quality across the evaluated workloads, while achieving up to $14.3\times$ TTFT speedup at 32K. In summary, our contributions are as follows:
\begin{enumerate}\itemsep2pt
\item We present Tail-Replay, to our knowledge the first hybrid prefix caching mechanism that enables unconstrained token-level prefix reuse without being constrained by recurrent-state checkpoints.
\item We introduce a replay-based state reconstruction mechanism that caches exact FA KV and rebuilds the  linear-attention state from a short, recent tail of the matched prefix.
\item We implement and evaluate Tail-Replay on three Gated DeltaNet-based hybrid LLMs, demonstrating high quality retention and substantial TTFT reduction.
\item We develop two quality-preserving optimizations---a tail-FFN skip and transfer/replay overlap---that further reduce replay overhead.
\end{enumerate}

\section{Preliminary and Related Work}

Hybrid LLMs interleave a small number of FA layers with linear-attention layers. In an FA layer, the reusable cache is token indexed: each token contributes its own KV pair. A linear-attention layer instead summarizes its prefix into a recurrent state $S_i$, updated token by token and exposed only through the final state after prefill.

For Gated DeltaNet~\cite{gateddeltanet}, the update takes the form
\begin{equation}
S_i = T_i S_{i-1} + \beta_i v_i k_i^{\top}, \qquad
T_i = \alpha_i\!\left(I - \beta_i k_i k_i^{\top}\right),
\end{equation}
where $i$ indexes the $i$-th token in the prefix, $k_i$ and $v_i$ are its key and value vectors, and $\alpha_i \in (0,1]$ is a learned gate. This gate progressively attenuates earlier information, while the rank-1 correction reshapes the current update. 

\paragraph{Related work.} Prefix caching systems for full-attention models reuse token-indexed KV caches, including RadixAttention~\cite{sglang}, Prompt Cache~\cite{promptcache}, CachedAttention~\cite{cachedattention}, and CacheBlend~\cite{cacheblend}. Position-independent caching (PIC) extends this idea to independently matched chunks~\cite{epic,prophetkv,hypic}, but still relies on token-addressable attention states. For hybrid LLMs, Marconi~\cite{marconi} and Sparse Prefix Caching~\cite{sparseprefix} retain recurrent checkpoints, while LinearKV~\cite{liu2026linearkvcachedstatesuffices} uses cached local linear states as initializers; Tail-Replay instead reconstructs the matched-prefix state from a short recent hidden suffix without storing recurrent checkpoints.

\section{Method}

\begin{figure*}[t]
\centering
\includegraphics[width=\textwidth]{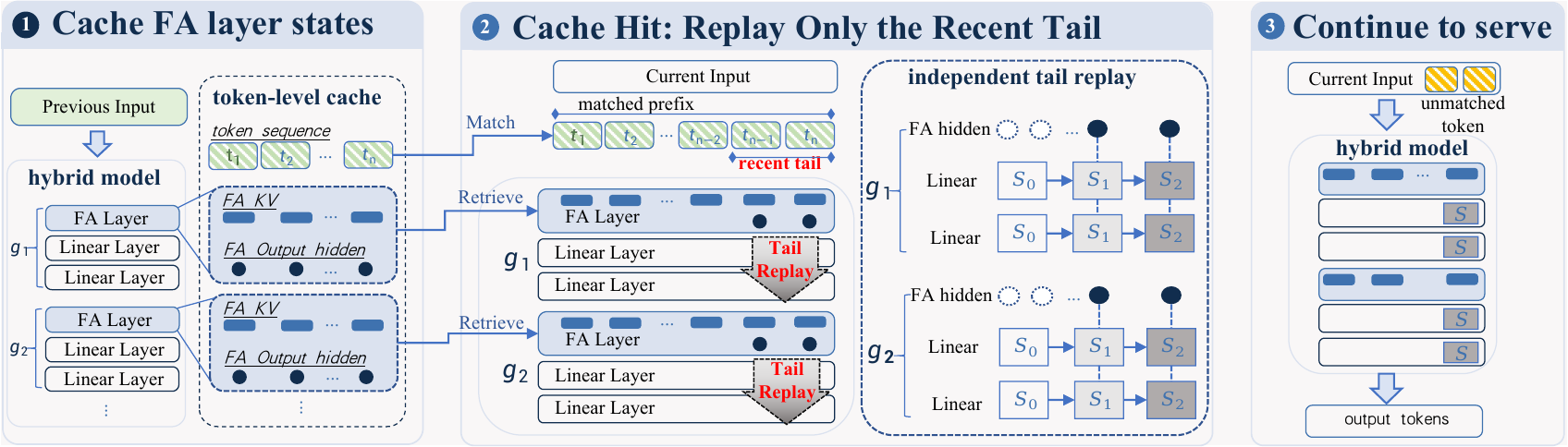}
\caption{Tail-Replay overview: cache FA KV and FA output hiddens, independently applies Tail-Replay to each linear-attention group on a cache hit, and then process the unmatched suffix.}
\label{fig:overview}
\end{figure*}

Figure~\ref{fig:overview} summarizes Tail-Replay. For each previous input, we retain only token-level states associated with the FA layers: the key/value pair and the corresponding FA output hidden for each token. Given a current request with a cache hit, the FA KV of the matched prefix is retrieved for direct reuse, while only a short, recent tail of the cached FA output hiddens is fetched as the replay input. Each group of linear-attention layers independently performs Tail-Replay from a zero-initialized state $S_0$ to reconstruct its recurrent state at the end of the matched prefix. The model then continues serving the unmatched tokens using the reconstructed linear-attention states and the retrieved FA KV.

\paragraph{Independent Tail-Replay.}

Tail-Replay approximates the recurrent state corresponding to a matched token prefix. Its accuracy is therefore governed by two factors: how much of the prefix is replayed and how faithfully the replay inputs match those used during the original prefill. The first factor is controlled by the replay ratio, for example by increasing the replayed tail from the latest $5\%$ to the latest $10\%$. The second factor motivates caching the output hidden of every FA layer.

Specifically, we cache the FA output hidden of each FA layer and partition the hybrid architecture into groups, each comprising one FA layer and the consecutive linear-attention layers that follow it. This ensures that the input hidden of the first linear-attention layer in every group exactly matches its value during the original prefill, thereby confining replay error within each group. Below, we describe one FA layer and its following linear-attention group; the same construction applies to every group.

Thus, for a previous input $P=(t_1,t_2,\ldots,t_n)$, the cache stores the token-level FA states $\{(K_i^{\mathrm{FA}},V_i^{\mathrm{FA}},h_i)\}_{i=1}^{n}$, where $(K_i^{\mathrm{FA}},V_i^{\mathrm{FA}})$ are the FA key/value vectors and $h_i$ is the FA output hidden. Given a cache hit covering the first $m$ tokens, we retrieve the corresponding FA KV and replay only the most recent $k=\lceil r m\rceil$ cached FA output hiddens. For one linear-attention group, we initialize the replay state at the beginning of the selected tail to zero, $\hat{S}_{m-k}=0$, and apply the GDN recurrence from Preliminary to the tail:
\begin{equation}
\hat{S}_{i}=T_i\hat{S}_{i-1}+\beta_i v_i^{\mathrm{LA}}\bigl(k_i^{\mathrm{LA}}\bigr)^{\top},\qquad i=m-k+1,\ldots,m,
\end{equation}
The cached FA output hiddens are fed through the group's linear-attention layers, which produce $k_i^{\mathrm{LA}}$, $v_i^{\mathrm{LA}}$, $T_i$, and $\beta_i$ at each replay step. The final state $\hat{S}_{m}$ approximates the recurrent state at the matched-prefix boundary. We perform this reconstruction independently for every group. Once all groups have been replayed, the model continues serving the current request by processing the unmatched suffix tokens with the retrieved FA KV and reconstructed linear-attention states.

\paragraph{Replay efficiency.} Since prefix caching aims to improve serving efficiency, we further reduce the overhead introduced by Tail-Replay with two optimizations. First, the FFN output at the end of a replayed group is not needed for state reconstruction: the next group starts from the exact FA output hidden of its following FA layer, which is already cached. We therefore omit this FFN during replay and compute only the linear-attention block of the group's final layer, which is sufficient to update the recurrent state. Second, because Tail-Replay does not depend on the cached FA KV, transferring the FA KV from host memory to the device can proceed concurrently with replay on a separate copy stream. The transfer is synchronized only before the query forward, allowing replay to hide most of the data-movement cost.

\section{Experiment}

\paragraph{Setup.} We evaluate three Gated DeltaNet-based hybrid LLMs---OLMo-Hybrid-7B, Qwen3.5-4B, and Qwen3.6-27B---on NVIDIA H100 GPUs using PyTorch 2.9.1.
\paragraph{End-to-end quality.}\label{sec:quality} We measure end-to-end quality on LongBench~\cite{longbench} and RULER~\cite{ruler}, comparing full prefill with Tail-Replay at $r\in\{5\%,10\%\}$. We also include a \zonly{} baseline, which reuses the matched FA KV but zeros the recurrent linear-attention states without replay, isolating the benefit of state reconstruction. Table~\ref{tab:quality} reports benchmark-level averages over all evaluated tasks and context lengths; each replay entry gives its absolute score followed by its percentage relative to full prefill. Tail-Replay retains 92.8--98.9\% of full-prefill quality on LongBench at $r{=}5\%$ and 93.9--98.1\% at $r{=}10\%$; on RULER, it retains 93.1--99.9\% and 96.7--99.9\%, respectively. 
\begin{table}[t]
\centering\scriptsize
\caption{Average quality scores within each benchmark and across all evaluated context lengths. LongBench averages its task-level scores (token-F1 or ROUGE-L), while RULER averages recall across its eight task--length cells. Replay entries show the absolute score followed by its percentage relative to full prefill (full $=100\%$); complete per-cell results are in Appendix~\ref{sec:quality_appendix}.}
\label{tab:quality}
\begin{tabular}{lrrrr|rrrr}
\toprule
& \multicolumn{4}{c|}{LongBench} & \multicolumn{4}{c}{RULER} \\
Model & Full & Zero & $r{=}5\%$ & $r{=}10\%$ & Full & Zero & $r{=}5\%$ & $r{=}10\%$ \\
\midrule
Qwen3.6-27B    & 0.426 & 0.313 & 0.410\ (96.2\%) & 0.417\ (97.8\%) & 0.987 & 0.299 & 0.985\ (99.8\%) & 0.986\ (99.9\%) \\
Qwen3.5-4B     & 0.374 & 0.271 & 0.369\ (98.9\%) & 0.367\ (98.1\%) & 0.960 & 0.415 & 0.959\ (99.9\%) & 0.958\ (99.7\%) \\
OLMo-Hybrid-7B & 0.317 & 0.159 & 0.294\ (92.8\%) & 0.297\ (93.9\%) & 0.812 & 0.215 & 0.756\ (93.1\%) & 0.785\ (96.7\%) \\
\bottomrule
\end{tabular}
\end{table}

\paragraph{Serving efficiency.} We evaluate serving efficiency with shared \texttt{narrativeqa} prefixes at nominal 8K, 16K, and 32K context lengths. We report TTFT from the start of cache transfer or replay to the first generated token. Table~\ref{tab:main} compares the full-prefill baseline with H2D-SER and H2D-OVL+skip at both replay budgets. The full-prefill cost grows substantially with context length, whereas the 5\% OVL+skip path remains comparatively stable: its speedup reaches $9.8\times$, $9.1\times$, and $14.3\times$ at 32K for OLMo-Hybrid-7B, Qwen3.5-4B, and Qwen3.6-27B, respectively. Relative to serialized H2D transfer, OVL+skip further reduces TTFT by 18--42\% at 32K. Increasing the replay budget to 10\% has little effect at shorter contexts but raises TTFT at 32K, where replay becomes the dominant cost.

\begin{table}[t] \centering\scriptsize \caption{TTFT (ms, mean over 20 timed repetitions). `H2D-SER' denotes serialized host-to-device transfer; `H2D-OVL+skip' denotes overlapped transfer with the FFN skip. The matched-prefix lengths are in tokens; speedups in parentheses are relative to same-row full prefill.} \label{tab:main} \begin{tabular}{llrrrrr} \toprule Model & Matched prefix length & \tf & \multicolumn{2}{c}{5\% replay} & \multicolumn{2}{c}{10\% replay} \\ \cmidrule(lr){4-5}\cmidrule(l){6-7} & & & H2D-SER & H2D-OVL+skip & H2D-SER & H2D-OVL+skip \\ \midrule OLMo-Hybrid-7B & 8{,}225  & 264.3  & 100.7 & 82.9\ (3.19$\times$) & 101.2 & 82.7\ (3.20$\times$) \\                & 16{,}417 & 533.7  & 122.3 & 83.5\ (6.39$\times$) & 121.7 & 85.6\ (6.24$\times$) \\                & 31{,}366 & 1068.3 & 158.8 & 108.8\ (9.82$\times$) & 191.7 & 112.5\ (9.50$\times$) \\ \addlinespace Qwen3.5-4B     & 8{,}226  & 191.6  & 81.3 & 75.2\ (2.55$\times$) & 82.0 & 75.9\ (2.53$\times$) \\                & 16{,}418 & 387.7  & 87.0 & 75.0\ (5.17$\times$) & 87.7 & 76.0\ (5.10$\times$) \\                & 31{,}847 & 786.4  & 105.8 & 86.2\ (9.12$\times$) & 133.9 & 108.1\ (7.27$\times$) \\ \addlinespace Qwen3.6-27B    & 8{,}226  & 879.6  & 166.6 & 154.3\ (5.70$\times$) & 167.4 & 154.7\ (5.69$\times$) \\                & 16{,}418 & 1797.2 & 187.7 & 161.6\ (11.12$\times$) & 260.8 & 218.1\ (8.24$\times$) \\                & 31{,}847 & 3605.2 & 311.9 & 251.8\ (14.32$\times$) & 453.9 & 371.3\ (9.71$\times$) \\ \bottomrule \end{tabular} \end{table}

\section{Conclusion}

Tail-Replay addresses the mismatch between token-level prefix sharing and recurrent state in hybrid LLMs. By caching exact FA KV and FA output hiddens, then independently replaying only a recent suffix for each linear-attention group, it enables flexible prefix reuse without recurrent-state checkpoints. Across three Gated DeltaNet-based hybrid models, short replay tails preserve most of full-prefill quality across the evaluated workloads while delivering substantial TTFT reductions at long contexts.

\newpage
\appendix
\section{Complete Quality Results}
\label{sec:quality_appendix}

Table~\ref{tab:quality_full} lists all quality cells used in Table~\ref{tab:quality}; values are absolute scores for full prefill, zero-only, and the two replay budgets.

\begin{table*}[h]
\centering\scriptsize
\caption{Complete per-cell quality results. LB denotes LongBench and R denotes RULER.}
\label{tab:quality_full}
\begin{tabular}{llrrrr}
\toprule
Model & Cell & Full & Zero-only & $r{=}5\%$ & $r{=}10\%$ \\ \midrule
Qwen3.5-4B & LB-narrativeqa@16K & .244 & .220 & .243 & .249 \\
Qwen3.5-4B & LB-narrativeqa@32K & .276 & .222 & .268 & .270 \\
Qwen3.5-4B & LB-narrativeqa@64K & .293 & .228 & .292 & .286 \\
Qwen3.5-4B & LB-hotpotqa@16K & .652 & .447 & .657 & .662 \\
Qwen3.5-4B & LB-qasper@16K & .487 & .355 & .482 & .461 \\
Qwen3.5-4B & LB-musique@16K & .439 & .223 & .422 & .419 \\
Qwen3.5-4B & LB-qmsum@16K & .225 & .202 & .223 & .221 \\
OLMo-Hybrid-7B & LB-narrativeqa@16K & .203 & .091 & .203 & .212 \\
OLMo-Hybrid-7B & LB-narrativeqa@31K & .219 & .098 & .214 & .217 \\
OLMo-Hybrid-7B & LB-hotpotqa@16K & .565 & .330 & .544 & .548 \\
OLMo-Hybrid-7B & LB-qasper@16K & .394 & .129 & .308 & .318 \\
OLMo-Hybrid-7B & LB-musique@16K & .305 & .107 & .282 & .273 \\
OLMo-Hybrid-7B & LB-qmsum@16K & .214 & .198 & .213 & .217 \\
Qwen3.6-27B & LB-narrativeqa@16K & .265 & .214 & .259 & .267 \\
Qwen3.6-27B & LB-narrativeqa@64K & .332 & .231 & .329 & .321 \\
Qwen3.6-27B & LB-hotpotqa@16K & .688 & .536 & .665 & .677 \\
Qwen3.6-27B & LB-qasper@16K & .521 & .344 & .481 & .484 \\
Qwen3.6-27B & LB-musique@16K & .518 & .346 & .498 & .519 \\
Qwen3.6-27B & LB-qmsum@16K & .233 & .208 & .227 & .231 \\
Qwen3.5-4B & R-cwe@8K & .998 & .169 & .998 & .998 \\
Qwen3.5-4B & R-cwe@16K & .994 & .215 & .994 & .994 \\
Qwen3.5-4B & R-qa1@8K & .860 & .405 & .845 & .850 \\
Qwen3.5-4B & R-qa1@16K & .830 & .330 & .835 & .820 \\
Qwen3.5-4B & R-niah@8K & 1.000 & .849 & 1.000 & 1.000 \\
Qwen3.5-4B & R-niah@16K & 1.000 & .786 & 1.000 & 1.000 \\
Qwen3.5-4B & R-vt@8K & 1.000 & .325 & 1.000 & 1.000 \\
Qwen3.5-4B & R-vt@16K & 1.000 & .244 & 1.000 & 1.000 \\
OLMo-Hybrid-7B & R-cwe@8K & .865 & .079 & .786 & .814 \\
OLMo-Hybrid-7B & R-cwe@16K & .544 & .023 & .534 & .558 \\
OLMo-Hybrid-7B & R-qa1@8K & .730 & .220 & .665 & .715 \\
OLMo-Hybrid-7B & R-qa1@16K & .725 & .190 & .645 & .665 \\
OLMo-Hybrid-7B & R-niah@8K & .983 & .624 & .904 & .929 \\
OLMo-Hybrid-7B & R-niah@16K & .968 & .580 & .895 & .924 \\
OLMo-Hybrid-7B & R-vt@8K & .885 & .006 & .847 & .868 \\
OLMo-Hybrid-7B & R-vt@16K & .795 & .000 & .773 & .805 \\
Qwen3.6-27B & R-cwe@8K & 1.000 & .092 & 1.000 & 1.000 \\
Qwen3.6-27B & R-cwe@16K & 1.000 & .061 & 1.000 & 1.000 \\
Qwen3.6-27B & R-qa1@8K & .895 & .405 & .880 & .885 \\
Qwen3.6-27B & R-qa1@16K & .875 & .375 & .880 & .885 \\
Qwen3.6-27B & R-niah@8K & 1.000 & .899 & 1.000 & 1.000 \\
Qwen3.6-27B & R-niah@16K & 1.000 & .881 & 1.000 & 1.000 \\
Qwen3.6-27B & R-vt@8K & 1.000 & .011 & 1.000 & 1.000 \\
Qwen3.6-27B & R-vt@16K & 1.000 & .060 & 1.000 & 1.000 \\
\bottomrule
\end{tabular}
\end{table*}

\label{endcontent}\typeout{CONTENT_ENDS_ON_PAGE:\thepage} \bibliographystyle{plainnat} \bibliography{references}

\end{document}